\documentclass[10pt,conference]{IEEEtran}
\usepackage{cite}
\usepackage{amsmath,amssymb,amsfonts}
\usepackage{algorithmic}
\usepackage{graphicx}
\usepackage{textcomp}
\usepackage{xcolor}
\usepackage{subfig}
\usepackage{caption}
\usepackage{epstopdf}
\usepackage{subfloat}
\usepackage{multirow}
\usepackage{float}

\usepackage{color}
\usepackage{booktabs}

\def\BibTeX{{\rm B\kern-.05em{\sc i\kern-.025em b}\kern-.08em
    T\kern-.1667em\lower.7ex\hbox{E}\kern-.125emX}}
\begin{document}
 \setlength{\columnsep}{0.25in}
 \captionsetup[figure]{name={Fig.},labelsep=period}
 \captionsetup[table]{name={TABLE.},labelsep=period}

\title{Exposing Deepfake with Pixel-wise Autoregressive and PPG Correlation from Faint Signals}
\author{Maoyu Mao\IEEEauthorrefmark{1}, Jun Yang\IEEEauthorrefmark{1}, \\
\IEEEauthorblockA{
\IEEEauthorrefmark{1}Dept. of Electronic and Information Engineering, Tongji University, Shanghai, China. Email: junyang@tongji.edu.cn\\}}

\maketitle

\begin{abstract}
Deepfake poses a serious threat to the reliability of judicial evidence and intellectual property protection. In spite of an urgent need for Deepfake identification, existing pixel-level detection methods are increasingly unable to resist the growing realism of fake videos and lack generalization. In this paper, we propose a scheme to expose Deepfake through faint signals hidden in face videos. This scheme extracts two types of minute information hidden between face pixels – photoplethysmography (PPG) features and auto-regressive (AR) features, which are used as the basis for forensics in the temporal and spatial domains, respectively. According to the principle of PPG, tracking the absorption of light by blood cells allows remote estimation of the temporal domains heart rate (HR) of face video, and irregular HR fluctuations can be seen as traces of tampering. On the other hand, AR coefficients are able to reflect the inter-pixel correlation, and can also reflect the traces of smoothing caused by up-sampling in the process of generating fake faces. Furthermore, the scheme combines asymmetric convolution block (ACBlock)-based improved densely connected networks (DenseNets) to achieve face video authenticity forensics. Its asymmetric convolutional structure enhances the robustness of network to the input feature image upside-down and left-right flipping, so that the sequence of feature stitching does not affect detection results. Simulation results show that our proposed scheme provides more accurate authenticity detection results on multiple deep forgery datasets and has better generalization compared to the benchmark strategy.
\end{abstract}
\let\thefootnote\relax\footnotetext{Corresponding author is Jun Yang. Email: junyang@tongji.edu.cn}

\section{Introduction}
Fake face video generated by advanced computer vision technology and deep learning technology poses a serious threat to the reliability of judicial evidence and intellectual property protection. Therefore, it is imperative to develop a universal Deepfake detection algorithm. Methods of Deepfake detection are divided into fake image detection and fake video detection\cite{Nguyen2019Deep}. In 2018, Pavel K et al. first conducted a fairly comprehensive evaluation of facial recognition methods for detecting Deepfake videos\cite{Korshunov2018DeepFakes}. They observed that, even though Visual Geometry Group (VGG) and FaceNet neural networks, the state of the art image classification models, are helpless for Deepfake video recognition. The compression of video results in the degradation of frame data, and the timing characteristics between frames may change, so image detection algorithms cannot be directly used for video detection.

Until recently, in order to resist an endless stream of Deepfake videos, researchers begin to develop datasets and countermeasure for such problems. In view of temporal structure of fake videos, the detection strategies for forgery videos can be divided into methods using temporal features across frames and visual artifacts within frame. Methods based on artifacts within frame usually combines image frame features combined with deep or shallow classifiers to achieve detection. A compact facial video forgery detection network named Mesonet proposed in\cite{Afchar2018MesoNet} utilizes deep neural network with a low number of layers to focus on the mesoscopic properties of images, as noise analysis cannot be used in compressed video detection with strong noise reduction while it is not easy to distinguish the authenticity of human face at semantic level. Five different detection systems evaluated thoroughly in\cite{Rossler2019FaceForensics} reveals that a linear stack of depthwise separable convolution layers with residual connections named XceptionNet providing the best results in both DeepFakes and FaceSwap manipulation methods. The authors also focus on Deepfakes, Face2Face and FaceSwap as outstanding of facial manipulations to build a general large-scale database. Additionally, background and foreground images containing fake faces are also statistically inconsistent during fusion, which is caused by inconsistent camera parameters, inconsistent lens illumination, compression in the generation algorithm and so on. When the synthetic face area is synthesized into the original image to generate Deepfake image, artifacts will be generated at the splicing place. Landmarks extracted in\cite{Lingzhi2020Face} display the mixed boundary information of the Deepfake images and describe the syncretic mask for showing face x-ray. Further, they adopt advanced high-resoultion nets (HRNets) to achieve two-class detection. Instead of making blind assumptions around specific facial manipulation algorithm and content, the methods relying on prior hidden information of videos can be widely effective. The basis of temporal features across frames methods is that the reconstruction process of Deepfake videos mostly relies on manipulation of frame-by-frame, therefore the temporal features between video frames cannot be thoroughly preserved. Intuitively, this type of methods can adopt deep recurrent neural networks (RNNs) for end-to-end detection\cite{Guera2018Deepfake, Sabir2019Recurrent}, while there is no strong theoretical support for directly inputting the original fake face video into network and pure pixel-level methods cannot resist the increasingly realistic fake videos. On the other hands, extracting temporal features by using the inconsistency of biological attributes on face provides a good solution. Errors caused by inconsistent head poses utilized in\cite{Yang2019Exposing} combines with shallow classifier support vector machine (SVM) to achieve convenient classification. The abnormal blinking state found in\cite{Li2018In} is modeled in combination with long-term recurrent convolutional neural networks (LRCNs) model to achieve Deepfake detection. Fidelity of lip movement in face videos evaluated in\cite{Agarwal2019Proceedings} combines sound synchronization analysis to judge the authenticity of video.

Fake face videos realized by deep algorithms such as GAN-based entire face synthesis or face indentify swap can learn pixel-based features to complete visual deception, but it cannot imitate the change rules of subtle signal, such as biological signal contained in all organisms. Non-electrical biological signals, including heartbeat, pulse, etc., fluctuate regularly, and can be used as fingerprints for authenticity detection as they will not be completely retained in the spatio-temporal domain of fake videos. As early as 2008, W. Verkruysse first proposes the technology that PPG signal related to heart rate can be extracted from face videos collected from camera so as to realize remote heart rate monitoring\cite{Verkruysse2008Remote}. This technology is named remote PPG (rPPG). Blood volume pulse from the facial regions extracted in\cite{Poh2010advancements} applies blind source separation (BBS) algorithm used in RGB color space and is subsequently utilized to quantify HR, respiratory rate, and HR variability. Significant motion renders vulnerability of previous algorithm. Difference in intensity of PPG signals in RGB color channel found in\cite{Verkruysse2008Remote} is used to calculate the linear combination and ratio of chrominance signals, and finally obtains chrominance-based rPPG, with root mean square error (RMSE) and standard deviation both a factor of 2 better than BSS-based algorithms. AR modelling and pole cancellation utilized in [19] cancels out aliased frequency components caused by artificial light flicker and constucts accurate HR spectrum from AR model. The rPPG technology is widely concerned and applied in the fields of medical imaging research\cite{Bruser2015Ambient}, emotion recognition\cite{Markova2019CLAS} and face anti-spoofing\cite{Liu2018Learning}. Lately, experimental verification in\cite{Ciftci2020FakeCatcher} shows that spatial coherence and temporal consistency of biological signals can be used to achieve Deepfake detection, with the video accuracy up to 94.65\%, which is the first time HR signals is mentioned in the field of deep forgery. On the basis of this confirmation, this team further proposes that spatio-temporal patterns of biological signals can be considered as representative projections of residuals\cite{CIftci2020How}. This novel scheme can realize deep forgery source detection according to the characteristics of generative models.

In this paper, we investigate the modeling of biological signal and its AR coefficient as implicit prior fingerprints and propose a deep forgery forensics scheme with the help of weak biological signals to address known or unknown facial manipulation detection. Since HR-related signal cannot be accurately reproduced by Deepfake generation network, and biological signals hidden in portrait video have continuous, stable and complex fluctuations, its spatio-temporal features can be extracted and applied to neural network to detect falseness in face videos. More specifically, in this scheme, the cheek part of the face image frame is selected as the ROI and is divided into sub-regions of same size. The PPG signal and its AR coefficient of the corresponding area of portrait videos are extracted one by one and modeled as two class of intuitive maps. We build two DenseNets that accept different feature representations (PPG map and AR map) of input and demonstrate the advantages of integrating our model detection. Simulation results show that the proposed deep forgery forensics model fusion can improve the detection accuracy, reduce the detection delay and enhance wide universality.

The main contributions of this paper can be summarized as:
\begin{itemize}
\item We explore novel remote biological signal and its variants as priori fingerprints for Deepfake forensics.
\item We formulate a bio-signal-based deep forgery forensics scheme to detect Deepfake face in portrait videos. Two types of maps are obtained by modeling the PPG signal and its AR coefficients, and then combined with deep neural networks to achieve two-class detection.
\item We utilize feature fusion technology to input the two types of feature representations into the network model and evaluate the performance. The proposed scheme improves the detection accuracy, reduces the detection delay and enhances wide universality.
\end{itemize}

The rest of this paper is structured as follows. We  review photoelectric plethysmography and AR coefficients in Section \uppercase\expandafter{\romannumeral2} and describe the feature map of proposed Deepfake forensic scheme in Section \uppercase\expandafter{\romannumeral3}. We propose a PPG-based Deepfake forensic scheme with fusion model \uppercase\expandafter{\romannumeral4} and provide simulation results in Section \uppercase\expandafter{\romannumeral5}. Conclusion is drawn in Section \uppercase\expandafter{\romannumeral6}.

\section{Related Work}

Despite the CNNs models are capable of generating fake images, these generated synthetic images are still detectable. What has greatly contributed to the popularity of Deepfake detection methods is the fact that the retained fingerprints generated by the CNNs model can be distinguished from real counterparts \cite{wang2020cnn}.
Creating images and videos are available today as it only rely on several real images or short videos as references. The Deepfake detection methods include fake iamge detection and fake video detection. Characteristics in videos are varied among different frames. Based on a obvious observation that the feature distribution of the real faces is crowded while that of the fake ones is uncrowded among domains but crowded in single domain, Jia et al\cite{jia2020single}. develops a single-side domain generalization framework for judging the feature space for real and fake faces. In particular, the generalization of the model is available by using the normalization function of feature and weight. Durall et al. \cite{durall2020watch} claim that one common limitation of the ordinary up-sampling models, e.g. well-known GAN models, is the incapability of reconstructing the same distribution of spectrum like that in true data. Therefore, they formulate a special spectrum regularizer for optimization both the fake images quality and training stability. Liu et al. \cite{liu2020global} believe that the texture information between fake and genuine faces is virtually different and there is a greater difference in global texture information between real faces and real ones.  Inspired by this view, the Gram-Net they proposed is shown to outperform the existing methods. Dang et al. \cite{dang2020detection} manifest that the attention module they proposed achieves good performance not only in learning the feature maps but also in distinguishing the genuine and fake faces. Meanwhile, the manipulated regions can be visualized. From a statistical point of view, due to the common systematic drawbacks on CNN-generated images, only utilize a single well-trained CNN, fake images generated by eleven different architectures can be picked out easily \cite{wang2020cnn}.

However, despite many potentials, these image detection methods are not suitable for detecting videos. Sabir et al.\cite{Sabir2019Recurrent} observe that the temporal coherence within the synthetic videos are weak, which can be calculated between two pictures before and after in a same video. The inconsistencies both on intra-frame and temporality has become an essential basis for judging true and false videos \cite{sun2018investigation}. As far as the method in general goes, the carefully configured combination of CNNs and long short term memory (LSTM) are widely used for detecting the spatial and temporal features, respectively \cite{donahue2015long,Guera2018Deepfake,chintha2020recurrent}.

The amount of light absorbed by blood varies naturally due to the flow of blood in the arteries, and vice versa. Therefore, we use the rPPG to detect such signals to identify genuine and fake faces, according to our method. Another observation is that even the generated Deepfake images are used by the "state-of-the-art" models, like GAN, the blurring process during upsampling still leaves traces in the images\cite{yu2019attributing}. Apart from this, the research on Deepfake detection using smoothing operations during up-sampling is still lacking. Kang et al. \cite{kang2013robust} propose to use an AR model after converting the image to a one-dimensional signal and achieved robustness in image median filtering detection. Subsequently, inspired by Kang's work, Yang et al. \cite{yang2018detecting} extend the AR model to two-dimensional space, directly enhancing its applicability in two-dimensional signal processing such as images. It is natural, to give us an inspiration that we can tell the difference between fake and true data from the correlation between pixels. Therefore, in this paper, AR is used to calculate the continuity between two adjacent pixels. When the correlation between pixels is low, it is very likely that the picture has been tampered with.
\begin{figure*}[!ht]
\centering
\centerline{\includegraphics[width=7.15in, height=2.95in]{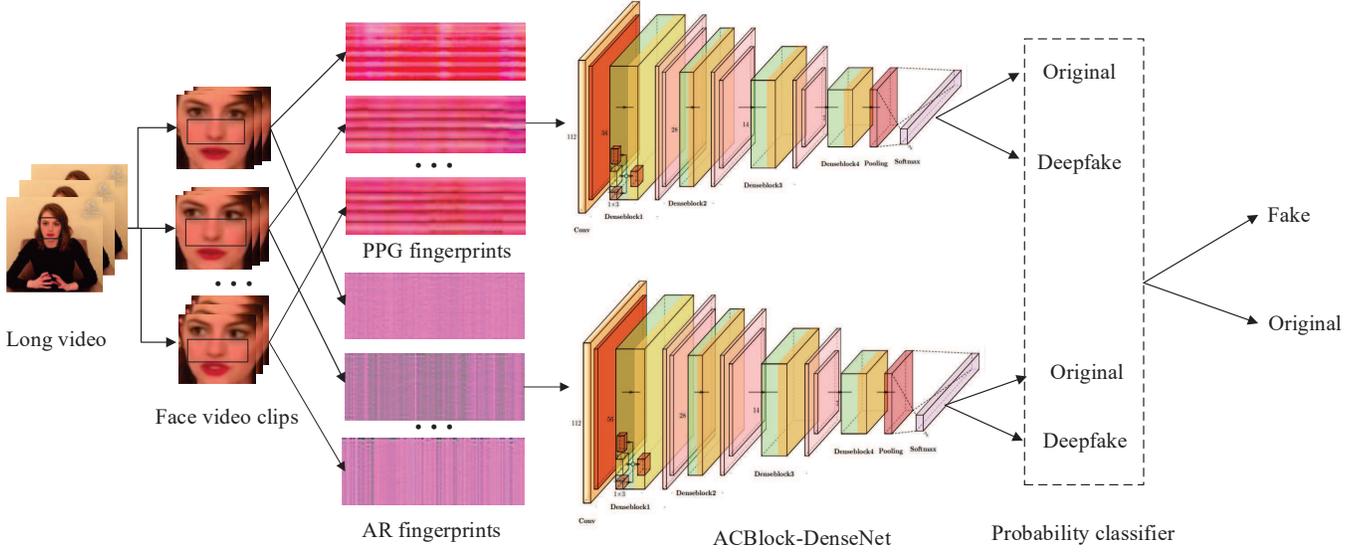}}
\caption{System model of proposed Deepfake detection scheme.}
\label{fig1}
\end{figure*}

\section{System Model}

\begin{table}[!t]
\caption{SUMMARY OF SYMBOLS AND NOTATIONS}
\begin{center}
\begin{tabular}{cc}
\hline
\textbf{\textit{Symbols}}& \textbf{\textit{Notations}}\\
\hline
$I(x,y,t)$ & Illumination intensity \\

$s$& Specular reflection \\

$\rho DC$& Direct current part \\

$\rho AC$& Alternating current part \\

$V_{i}(x,y,t)$& Intensity of reflected light\\

$i$& Color channel\\

$X_{s}(t), Y_{s}(t)$& The orthogonal chrominance signal\\

$C(t)$& Chrominance-PPG signal\\

$L(t)$& AR model \\

$p$&  Order of AR \\

$\varphi_{i}$& The autocorrelation coefficient \\
$c$& Constant \\
$\epsilon_{i}$& White noise \\

$L(x,y)$ & Pixel coordinate \\

$C(n)$ & PPG signal \\
$ROI$ & Region of interest\\

$\phi$ & AR coefficient matrix \\

$n$ & Frame with a fixed length \\

\hline

\end{tabular}
\label{tab1}
\end{center}
\end{table}

The system model of the proposed Deepfake detection scheme is shown in Fig.~\ref{fig1}. First, the face video is divided into multiple face image frame segments. The cheek region is selected as the ROI to generate PPG and AR fingerprint images. Two types of fingerprints are put into the DenseNet modified by ACBlock for training. Finally, model fusion is performed to achieve segment-level and video-level authenticity detection.

\subsection{Biological Signal-PPG signal for HR}

Photoplethysmography is an optical technique that monitors various vital signs, such as heart rate, which exploits photoelectric sensors to detect and record the light absorption or reflection intensity variations of the human skin due to the blood volume variations during the cardiac cycle, thereby achieve the remote extraction of biological signals.

Extracting frame sequence of video $t,t=1,2,3...$ recorded by the light sensitive sensor in the camera, the intensity of the reflected light from the skin surface recorded by the video can be expressed as
\begin{equation}
V(x,y,t) =I(x,y,t)R
\label{eq0}\,,
\end{equation}
where $I(x,y,t)$ is the illumination intensity of the light source at the pixel coordinates $(x,y)$ of the video frames and $R$ is the reflectance of the skin surface. Light produces specular and diffuse reflections on the facial skin.
Therefore, the reflectance $R$ can be further decomposed as
\begin{equation}
R =s+\rho,
\label{eq00}\,
\end{equation}
where $s$ and $\rho$ denote specular reflection and diffuse reflection, respectively. Analyzing the interaction of the structural level of the skin with light, the diffuse reflection $\rho$ is divided into a direct current part $\rho_{DC}$ and an alternating current part $\rho_{AC}$, i.e.
\begin{equation}
\rho(t) =\rho_{DC}+\rho_{AC}
\label{eq000}\,.
\end{equation}

For the selected region of interest (ROI), the intensity of reflected light $V_{i}(x,y,t)$ for each color channel $i\in\{R,G,B\}$ can be modeled as
\begin{equation}
V_{i}(x,y,t) =I_{i}(x,y,t)(s+ \rho_{i_{DC}} +\rho_{i_{AC}}) \label{eq1}\,,
\end{equation}
where $s$ is is identical for all the color channels, $\rho_{i_{DC}}$ is the stationary part of the reflection coefficient of the skin in the color channel $i$, while $\rho_{i_{AC}}$ is the zero-mean time-varying physiological waveform attributed to cardiac synchronous changes in the blood volume with heart rate.

To reduce motion artifacts and other noise effects parallel to the imaging plane, a group of pixels $M\times N$ in ROI is selected for averaging pooling. Then the chrominance signal of each color channel is
\begin{equation}
C_{i}(t)=\frac{\sum_{y=0}^{N-1}\sum_{x=0}^{M-1}V_{i}( x,y,t)}{M\times N}\,, \: \: \: \:  i\in \{R,G,B\}\,.
\label{eq2}
\end{equation}

In a white-light illumination environment, due to differences in the mapping of skin tones of different individuals, we correct the chrominance signal $C_{i}(t)$ with the help of normalization of the differences as
\begin{equation}
C_{i_{n}}(t)=\frac{C_{i}(t)}{\sigma}\,,
\end{equation}
where $\sigma$ is defined as $\sqrt{C_{R}(t)^2+C_{G}(t)^2+C_{B}(t)^2}$. In a white lighting environment, the standard skin colour vector $\sigma$ is noted as $[0.7682, 0.5121, 0.3841]$. Under white light illumination, the angles between different skin colors and the standard skin color vectors are approximately equally distributed in colour space. Therefore, the corrected chrominance signal can be expressed as
\begin{equation}
\begin{aligned}
&[C_{R_{n}}(t), C_{G_{n}}(t), C_{B_{n}}(t)]\\
=&[0.7682C_{R}(t), 0.5121C_{G}(t),0.3841C_{B}(t)].
\end{aligned}
\end{equation}

Two orthogonal chrominance signals are utilized to eliminate specular reflection component in \eqref{eq1}, i.e.,
\begin{equation}
\begin{aligned}
X_{s}(t)&=\frac{C_{R_{n}}(t)-C_{G_{n}}(t)}{0.7682-0.5121}=3C_{R}(t)-2C_{G}(t) \\
Y_{s}(t)&=\frac{0.5C_{R_{n}}(t)+0.5C_{G_{n}}(t)-C_{B_{n}}(t)}{0.7682+0.5121-0.7682}\\
&=1.5C_{R}(t)+C_{G}(t)-1.5C_{B}(t)\,.
\end{aligned}
\end{equation}

When the skin surface moves with respect to the light source, the illumination intensity in \eqref{eq1} will change and affect the chrominance intensity. However, such intensity modulations are equal for all channels, hence the influence of this motion can be eliminated by calculating the ratio of two filtered orthogonal signals. This ratio can be represented as
\begin{equation}
S(t)=\frac{X_{s}(t)}{Y_{s}(t)}-1\,.
\end{equation}
According to Taylor expansion, the approximate ratio $\hat{S}(t)$ is
\begin{equation}
\begin{aligned}
\hat{S}(t)&\approx X_{s}(t)-Y_{s}(t)\\&=1.5C_{R_{f}}(t)-3C_{G_{f}}(t)+1.5C_{B_{f}}(t)\,,
\end{aligned}
\end{equation}
where $C_{i_{f}}(t),i\in\{R,G,B\}$ is the bandpassed filtered version of $C_{i}(t)$.

Finally, the difference between the two frames of signals is used to reduce the influence of pigment absorption on diffuse reflection, and a PPG signal based on chromaticity changes is obtained as
\begin{equation}
C(t)=\hat{S}(t+1)-\hat{S}(t)\,.
\label{eq008}
\end{equation}
$C(t)$ is the chrominance-PPG (C-PPG) signal containing blood volume variation information. If needed, based on the multiple signal classification (MUSIC) algorithm, an estimate value of heart rate is obtained by calculating the pseudo-spectral peak in signal subspace.

\begin{figure*}[ht]       
\centering
\subfloat[Process of generating PPG fingerprints.]{\includegraphics[width=7.0in, height=1.35in]{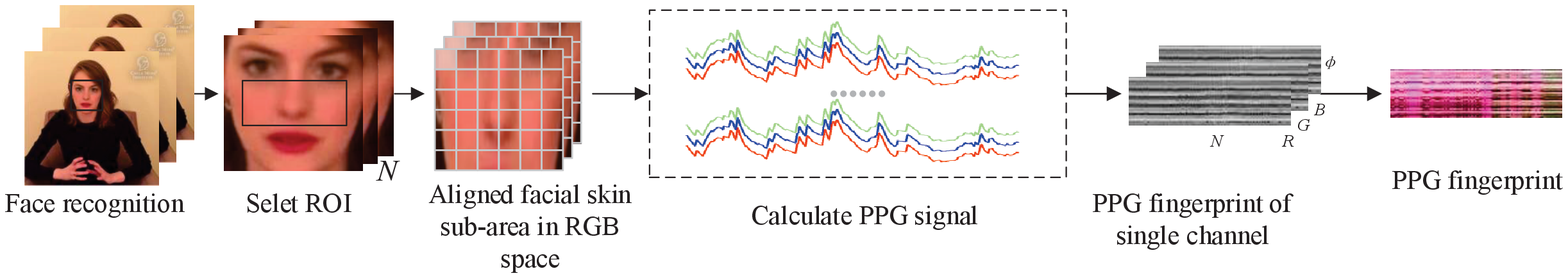}}\vspace{-0.1cm}
\subfloat[Process of generating AR fingerprints.]{\includegraphics[width=7.0in, height=1.25in]{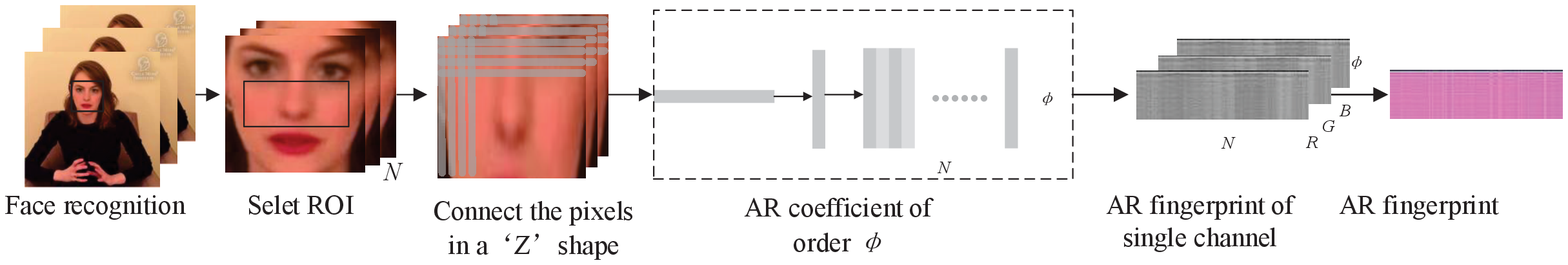}}
\caption{Process of generating fingerprints. With $\phi=36$ and $n=128$, fingerprints of size $36\times 128$ are obtained.}
\label{ppg_fingerprint}
\end{figure*}

\subsection{Autoregressive Model}

For a stationary non-white noise sequence, a linear model is usually established to fit the trend of the sequence in statistics, and the useful information in the sequence is extracted by this. The autoregressive(AR) model is a representation of a type of random process, which specifies that the output variable depends linearly on its own previous values and on a stochastic term, i.e.,
\begin{equation}
L_{t}=c+\sum_{i=1}^{p}\varphi_{i}L_{t-i}+\epsilon_{i}\,.
\end{equation}
where $p$ is the order of AR model, $\varphi_{i}$ is the autocorrelation coefficient, $c$ is a constant, and $\epsilon_{i}$ is white noise.

AR model can be used as a linear prediction model, which predicts the current sample based on evaluating the correlation between current sample and previous sample. In the established AR model, the calculated correlation coefficient is used to evaluate the correlation between current sample and previous sample. Based on camera imaging principles, we assume that pixel points are generated one by one. Pixels in real facial image have a strong correlation, which can be modeled as a stable AR processing model
\begin{equation}
L(x,y)=c+\sum_{i=1}^{p}\varphi_{i}L(x-i,y)+\epsilon_{i}\,,
\end{equation}
or
\begin{equation}
L(x,y)=c+\sum_{i=1}^{p}\varphi_{i}L(x,y-i)+\epsilon_{i}\,,
\label{eq009}
\end{equation}
where $L(x,y)$ is the pixel coordinate of facial images.

In the established model, the calculated correlation coefficient is used to evaluate the correlation between current sample and previous sample.

\subsection{Fingerprints from PPG and AR}

\begin{figure*}[!ht]
\centering
\centerline{\includegraphics[width=7.15in, height=1.77in]{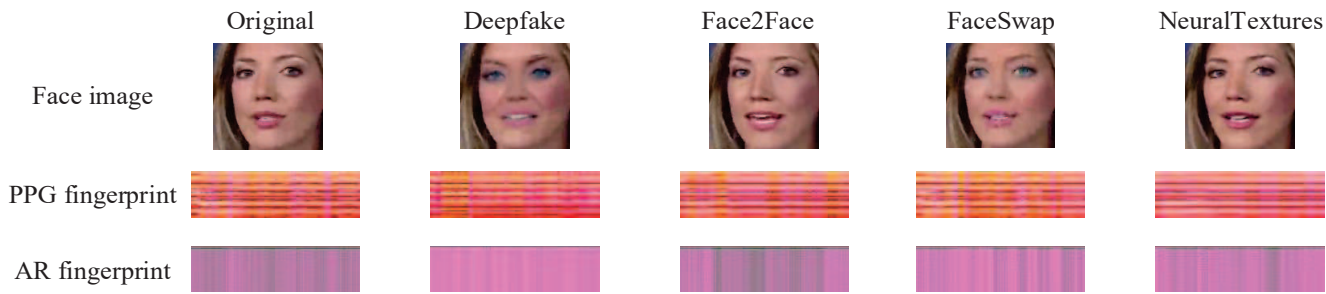}}
\caption{Examples of face images and their PPG fingerprints, AR fingerprints. Real faces and faces generated by Deepfake, Face2Face, FaceSwap, NeuralTextures models (top) are extracted to PPG fingerprints (middle) and AR fingerprints (bottom) with our scheme.}
\label{map}
\end{figure*}

Time and space information hidden in the fake face video will change. PPG is good at detecting changes in time domain, and AR has the ability to detect the relationship in spatial domain. Therefore, this article innovatively combines the two signals to obtain fusion fingerprint information based on the time-spatial domain.

Pure PPG signals have steady, periodic fluctuations in the heart rate range. Fake face video generated on a frame-by-frame basis disrupts the quasi-periodicity of the PPG signal, resulting in a jump in its instantaneous phase. Therefore, the abnormal changes in the PPG signal can be seen as fingerprints of a Deepfake face. The process of generating a PPG fingerprint is shown in Fig.~\ref{ppg_fingerprint} (a). In particular:

1) Extract a collection of face images from a video of frame length $t$ using face detector.

2) Select the pixels enclosed by four pixel coordinates of the cheek region as the region of interest (ROI) $R_o$ for feature extraction. The capillaries in the cheek area are abundant and generally not obscured by hair, accessories, etc.

3) Construct point-line relationship by Triangulation and stretch $R_o$ into a regular rectangle $R_n$ by means of affine transformation. Divide each frame of $R_n$ uniformly into 36 equal area non-overlapping sub-regions:$R{_n}(1), R{_n}(2), \ldots, R{_n}(i), \ldots, R{_n}(36)$.

4) Align facial skin ROI $R{_n}(i)$ in color space RGB. Separate three dimensional chromaticity signals of RGB for each subregion.  According to Equation \eqref{eq008}, calculate the PPG signal $C(n)$ of each sub-region over a range of frames of fixed length $n{(n<t)}$, and normalize it to the range of $[0-255]$. Replace B channel that contains the least obvious HR information with the PPG signal $C(n)$.

5) Calculate PPG signal value of length $n$ from each sub-region. Construct a matrix with 36 rows and n columns of signal values of length n in 36 sub-regions. In the same way, the R and G channels are reconstructed to obtain three $36\times n$ grayscale images. Finally, merge them into a color image as a PPG fingerprint of a video clip.

Fig.~\ref{map} shows sample facial images generated by original and each deepfake method (top), and an example of an original PPG fingerprint and a deep forged PPG fingerprint generated from the same window (middle).

Each row or column of pixels on the picture has a certain correlation, so it can be modeled as an AR model. The order of the AR can be used as a one-dimensional AR signal to describe the relevant information between pixel points. The process of generating fake face images, such as GAN-based face generation, will include changes in frequency features caused by smoothing operations such as up-sampling, and the correlation between pixel points also changes. The process of generating an AR fingerprint is shown in Fig.~\ref{ppg_fingerprint} (b). In particular:

1) Extract a collection of face images from a video of frame length $t$ using face detector.

2) Select the pixels enclosed by four pixel coordinates of the cheek region as the region of interest (ROI) $R_o$ for feature extraction. The capillaries in the cheek area are abundant and generally not obscured by hair, accessories, etc.

3) Construct point-line relationship by Triangulation and stretch $R_o$ into a regular rectangle $R_n$ by means of affine transformation.

4) Traverse the pixel values of rows $1, 3, 5,\ldots,2n-1$ in the order from left to right, and traverse the pixel values of rows $2, 4,\ldots, 2n$ in the order from right to left when building an AR model. They are connected row by row. According to camera imaging principles, pixel dots are generated one by one in $\rightleftarrows$ sequence, i.e. the pixels at the end of an odd numbered row are more correlated with the pixels at the end of the next even numbered row. Same thing with column traversal. Finally, the grey values in both directions will be joined to give a one-dimensional sequence of pixel values.

5) Set the order of AR to the number of blocks in PPG sub-region, i.e. 36 orders. Calculate the one-dimensional model coefficient matrix $\phi$ of the AR model, according to Equation \eqref{eq009}.

6) Calculate the AR coefficient matrix $\phi$ of each frame within the range of the number of frames with a fixed length of $n (n<t)$, and construct a $36\times n$ gray-scale image. Similarly, three 36×n grey-scale images based on the RGB colour space are obtained.Finally, merge the three into a color image as an AR fingerprint of a video clip.

Fig.~\ref{map} shows an example of an original AR fingerprint and a deep forged AR fingerprint generated from the same window (bottom).

\subsection{Authenticity discrimination with ACNet}

Instead of the traditional neural network classifier,  a novel convolutional layer structure, derived from ACNet, is utilized on top of DenseNet to identify deep face forgery. In general networks, $3\times 3$ convolution kernels are widely used. ACNet splits the $3\times 3$ convolution into asymmetric convolution in order to improve the efficiency and robustness of feature extraction.

In training phase, we replace each $3\times 3$ convolutional kernel with an AC block (ACB) containing $3\times 3$, $1\times 3$ and $3\times 1$ convolutional kernels, perform the convolutional operations separately and summed up their final results. Once the convolutional kernels have been fused, the same $3\times 3$ structure is used as the normal convolutional kernels in the testing phase, without adding additional computation.

The asymmetric convolution structure improves the robustness of the network to the up-and-down and left-right flipping of the input feature images, which has the advantage that the detection results are not affected regardless of the order in which our input fingerprint images are stitched together.

An overview of the network architecture for Deepfake detection is shown in Fig.~\ref{fig1}. The structure of DenseNet121 model includes $1$ large scale convolution layer firstly, $1$ pooling layers, $4$ Dense blocks, $3$ transition layers, $1$ full connection layer, input layer and output layer. Replacing the $3\times 3$ convolution kernels in the network with the structure of the ACB enhances the robustness of image inversion up and down and left and right, thereby eliminating the effect of the different stitching order of PPG or AR features.

\section{Simulation Results}

\begin{table*}[ht]
\caption{Qualitative comparison: authenticity detection accuracy (\%) of different depth forgery detection schemes.}
\begin{center}
\begin{tabular}{cccccccc}
\toprule
\multicolumn{2}{c}{\textbf{Manipulations}}          & \textbf{Xception}   & \textbf{MesoNet}    & \multicolumn{2}{c}{\textbf{\cite{Ciftci2020FakeCatcher}}} & \multicolumn{2}{c}{\textbf{Ours}} \\ \midrule
  & & video acc. & video acc. & segment acc.    & video acc.    & segment acc.  & video acc. \\ \cline{1-2}
\multirow{5}*{FaceForensics++}      & Deepfakes      & 94.28      & 89.52      & 90.76           & 94.87         & 93.71       & \textbf{96.44}      \\
                                 & Face2Face      & 91.56      & 84.44      & 93.12           & 96            & 87.26       & 90.50      \\
 & FaceSwap       & 93.70      & 83.56      & 94.22           & 95.75         & 95.45       & \textbf{95.95}      \\
                                  & NeuralTextures & 82.11      & 75.74      & -               & -             & 77.66       & \textbf{82.97}      \\
                                  & All            &   -   &   -      & 91.62           & 94.65         &  92.93  &     \textbf{96.13}  \\ \cline{1-3}
\multicolumn{2}{c}{FaceForensics}                 & 87.81      & 82.13      & 88.97           & 90.66         & 91.49       & \textbf{94.01}      \\ \bottomrule
\end{tabular}
\end{center}
\label{table2}
\end{table*}

\begin{table}[!ht]
\caption{Ablation experiments: detection accuracy (\%) of original RGB signal with ACNet, removing AR fingerprint (i.e., PPG fingerprint with ACNet), removing PPG fingerprint, removing ACNet (i.e., AR fingerprint and PPG fingerprint with CNN), and our proposed scheme.}
\begin{center}
\begin{tabular}{cccccc}
\toprule
\textbf{Manipulations}   & \textbf{RGB+ACNet} & \textbf{-AR} & \textbf{-PPG} & \textbf{-ACNet} & \textbf{Ours}  \\ \midrule
Deepfakes      & 85.32     & 96.01     & 93.20    & 92.50      & 96.44 \\
Face2Face      & 76.61     & 88.93     & 82.17    & 89.40      & 90.50 \\
FaceSwap       & 80.01     & 92.29     & 94.58    & 92.70      & 95.95 \\
NeuralTextures & 72.57     & 90.04     & 82.26    & 79.59      & 82.97 \\
All            &   78.75  & 93.38  & 93.65     &  93.15   &
96.13\\ \bottomrule
\end{tabular}
\end{center}
\label{table3}
\end{table}

The system is implemented in python, using the dlib library for face detection, OpenCV for image processing and Keras for neural network implementation. Most of the networks are trained and tested on 4 Tesla V100 GPUs, with short training times. The most computationally intensive part of the system is extracting PPG and AR fingerprints from large datasets, requiring two processes per video.

\begin{figure}[h]
\centering
\centerline{\includegraphics[width=3.55in, height=2.57in]{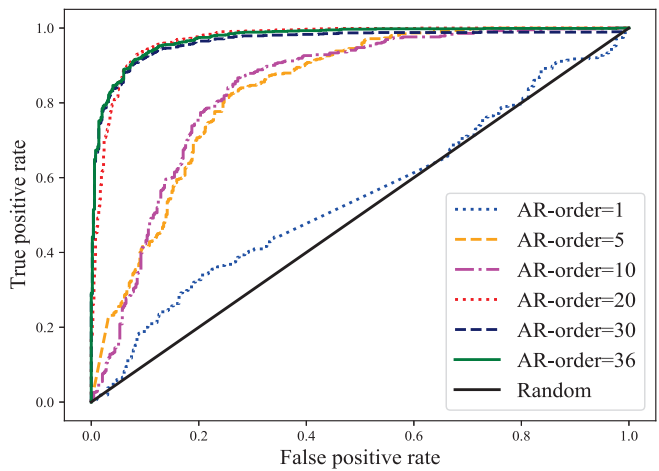}}
\caption{ROC curves and AUC values with different AR orders.}
\label{ar_order}
\end{figure}

FaceForensics++ is a facial forgery dataset that enables researchers to train deep learning-based methods in a supervised manner, which includes $4000$ fake face videos from $1000$ real vdieos, separately generated by four submethods: DeepFake (DF), Face2Face (F2F), FaceSwap (FS) and NeuralTextures (NT). Celeb-DF dataset contains $590$ real and $5639$ DeepFake synthesized videos with subjects of different ages, ethic groups and genders.

Simulations are performed to evaluate the performance of our methods with fixed frame length $n=128$, each fingerprint image is 128 in length and 36 in width. In the simulations, $70\%$ of the original fingerprints figures and Deepfake-based fignerprints figures are randomly chosen to train the network, and $30\%$ are randomly selected to be the testing set. In order to calculate the video detection accuracy, when randomly selecting, it is necessary to ensure that the PPG and AR fingerprint images extracted from the same video are either all in training set or all in test set.

\subsection{Order analysis of AR}
ROC (receiver operating characteristic) curves remain credibly evaluated on the unbalanced data set because they are not affected by the proportion of the category distribution. On the full dataset of FaceForensics++, ROC curves with AR orders of 1, 5, 10, 20, 30, and 36 are shown in Fig.~\ref{ar_order}. We can see that the AUC rises until the 20th order, after which the increase in order no longer plays a role in the AUC (area under curve) of the model. Even when the order goes to 30
or 36, the AUC is stable and saturated at around 0.98. For example, when the order is 5, the AUC is only 0.8841; when the order is 20, the AUC reaches 0.98, and when the order is 36, the AUC is still stable at 0.9835. It should be noted that the AUC rate has been stabilized, and increasing the order will not improve the performance of our model. In order to generate an image with the same size as the PPG fingerprint, the AR order is selected as 36 to calculate the AR fingerprint.

\subsection{Quantitative comparison}
Quantitative comparison between the proposed method and the method based on basic neural network and other weak signal methods is shown in Table.~\ref{table2}.
Our method calculates video-level accuracy, while calculating segment accuracy using the number of segment frames n=128. This is similar to the original PPG-based method proposed by \cite{Ciftci2020FakeCatcher}.

It can be seen from the table that when the FaceForensic++ dataset is selected as the experimental data, the detection accuracy is improved from 70.47\% to 96.13\% compared with the basic algorithm based on neural network. Compared with the method \cite{Ciftci2020FakeCatcher}, which is also based on weak hidden information, the segment detection accuracy has increased from 91.61\% to 92.93\%, and the video detection accuracy has increased from 94.65\% to 96.13\%. Similarly, when the original FaceForensic is selected as the experimental data, compared with the basic network algorithm, the video detection accuracy is improved by 11.88\%. Compared with the comparison algorithm \cite{Ciftci2020FakeCatcher}, the segment accuracy is increased by 2.52\%, and the video accuracy is increased by 3.35\%. Because we not only improved the method of generating PPG fingerprints, but also used the AR model to consider the correlation between pixels in the frame. Furthermore, we choose ACNet as the classifier, which is more robust in feature arrangement, for experiments.

For different generative models, the four models in FaceForensic++ are quantitatively compared. Video detection accuracy is higher with the proposed methods than with the basic network model algorithm. For example, on the fake face video data based on the Deepfakes generation model, the video detection accuracy has increased from 89.52\% to 96.44\% using MesoNet. In addition, using Xception to detect fake faces based on the FaceSwap generative model, the video detection accuracy increased from 93.70\% to 95.95\%. Compared with the comparison algorithm \cite{Ciftci2020FakeCatcher}, in addition to the lower accuracy of fake face video detection based on the Face2Face generative model, the proposed solutions all show better detection performance.

\subsection{Ablation experiments}
In order to demonstrate the effectiveness of the various components of the proposed scheme, ablation experiments with different component combinations are shown in Table.~\ref{table3}. In summary, compared with the ablation experiment that removes any components, the proposed scheme shows higher video detection accuracy in different fake face generation models and overall authenticity detection. Specifically, comparing the last two columns in the table, the overall accuracy is slightly improved. This reflects the improvement of the robustness of the ACNet asymmetric convolution structure to the selected feature fingerprints upside-down and left-right upside down, eliminates the influence of the feature arrangement order, and enhances the accuracy of model detection. Comparing the second and third columns in the table, the overall accuracy rate has increased from 81.89\% to 93.39\%, confirming the main contribution of the reference model \cite{Ciftci2020FakeCatcher}. That is, the heart rate signal in the real video has not been fully reproduced in the deep fake video, so it can be used for authenticity detection. Taking the last column as a reference, the overall accuracy of the third and fourth columns has increased by 2.74\% and 2.48\% respectively. This respectively reflects the gain effect of fingerprint features based on AR coefficients on the detection of the proposed authenticity detection scheme in the space domain, and fingerprint features based on PPG signals on the detection of the proposed authenticity detection scheme in the time domain. Neither of the first two columns in the table can well judge the authenticity of the face video. For example, in an experiment based only on RGB channels without any feature preprocessing, the video detection accuracy can only reach about 80\%; in an experiment that detects directly through a complete frame, the authenticity detection method is almost ineffective. The experimental results re-proven two points of view:
(1) The feature preprocessing based on the sparse AR sparseness of the heart rate signal is the key factor for authenticity detection.
(2) In the fake face detection experiment, it is necessary to extract the face in advance instead of the complete video frame to improve the detection performance.

\subsection{Cross-model and cross-domain evaluation}
\begin{table}
\caption{Detection accuracy (\%) of cross Deepfake generation model and cross fake face video data domain.}
\begin{center}
\begin{tabular}{cccccc}
\toprule
\textbf{Train set} & \textbf{Test set} & \textbf{Video acc.} &    \\ \midrule
FF++ - Deepfakes     & Deepfakes   & 95.75   \\
FF++ - Face2Face      & Face2Face  & 92.47 \\
FF++ - FaceSwap       &FaceSwap    & 96.15\\
FF++ - NeuralTextures & NeuralTextures & 79.25 \\
\cline{1-2}
FF++                   & FF          &     90.66    \\
FF++            & Celeb-DF        & 86.57 \\
\bottomrule
\end{tabular}
\label{table_4}
\end{center}
\end{table}
The cross-depth fake face generation model and the cross-fake face video data domain authenticity detection performance are shown in Table.~\ref{table_4}. In the cross-method evaluation of the four generative models in FaceForensic++, except for NeuralTextures, all have obtained high video detection accuracy. For example, when the model is trained with the remaining categories except FaceSwap, and then the fingerprint data of the FaceSwap generated model is used for detection, the video detection accuracy can be achieved by the accuracy of $96.15\%$. The cross-model evaluation and detection performance of the NeuralTextures generative model in the table is only $79.25\%$, because it is a generative model that is fundamentally different from other generative models.

In addition, the cross-domain evaluation results of the FaceForensic++ dataset and its predecessor, FaceForensic, and the more authentic Celeb-DF are shown in the last two rows of Table.~\ref{table_4}. Detected on the more streamlined original FaceForensic, the performance is very good, indicating that as long as the training data is sufficiently representative, the model can make correct predictions. Detecting on a more realistic Celeb-DF also obtains better detection performance, which verifies that the proposed scheme has generalization and can be widely used in various real-world scenarios.



\section{Conclusion}
In this paper, we propose a Deepfake detection scheme based on PPG signals and AR coefficients. In this scheme, the PPG signal is used to reflect the remote HR, and its fluctuation is regarded as temporal-domain features; the AR model is used to reflect the inter-pixel correlation, and the changes in its coefficients are treated as spatial-domain features. Furthermore, this scheme uses ACBlock-based DenseNet to evaluate the temporal-spatial fingerprints generated by two features, thereby realizing automatic authenticity detection. According to the simulation results, the proposed forensic scheme improves the accuracy and generalization of face video authenticity detection. For example, the simulation results show that on the FaceForensic++ dataset, the scheme improves the total accuracy by $25.66\%$ compared to baseline network, and $1.48\%$ compared to the baseline strategy in \cite{Ciftci2020FakeCatcher}. In addition, in cross-domain evaluation, the model trained on the FaceForensic++ dataset by the proposed scheme is able to achieve an accuracy of $86.57\%$ when tested on Celeb-DF dataset. The proposed false face forensics scheme can effectively improve the authenticity detection performance to achieve the reliability of judicial forensics and intellectual property protection. Future research will be devoted to explore the possibility of more faint information for face authenticity detection.

\bibliography{conference_041818}
\bibliographystyle{ieeetr}





\end{document}